\documentclass[11pt,letterpaper]{article}

\usepackage[T1]{fontenc}
\usepackage[utf8]{inputenc}
\usepackage{geometry}
\usepackage{setspace}

\usepackage[
    backend=biber,
    style=chem-angew,
    sorting=none,
    natbib=true
]{biblatex}
\usepackage{graphicx}
\usepackage{float}
\newfloat{scheme}{htbp}{los}
\floatname{scheme}{Scheme}
\floatname{chart}{Chart}
\newfloat{graph}{htbp}{loh}

\usepackage{chemformula}
\usepackage[version = 4]{mhchem}
\usepackage{amsmath,mathtools}
\usepackage{newtxtext,newtxmath}
\usepackage{booktabs,multirow,array}
\usepackage{xcolor}
\usepackage{microtype}
\usepackage{nicefrac}
\usepackage{enumitem}
\usepackage{caption}
\usepackage{url}
\usepackage[
    colorlinks=true,
    citecolor=blue,
    linkcolor=blue,
    urlcolor=blue
]{hyperref}

\graphicspath{{figures/}}
\newcommand{\method}{MV-FREX}
\newcommand{\cmnpd}{CMNPD}

\DeclareMathOperator{\TopK}{TopK}
\newtheorem{proposition}{Proposition}

\usepackage{authblk}
\author[1]{Jiacheng Zheng*}
\author[2]{Chang Guo}
\author[3]{Zixuan Wang}
\author[4]{Xinyu Liu}
\author[1]{Hao Chen}
\affil[1]{Marine College, Shandong University, No. 180 West Wenhua Road, Weihai, Shandong, China}
\affil[2]{Department of Mathematics, Faculty of Mathematical \& Physical Sciences, University College London, London, United Kingdom}
\affil[3]{School of Life Science and Technology, Harbin Institute of Technology, Harbin, China}
\affil[4]{School of International Trade and Economics, University of International Business and Economics, Beijing, China}

\title{Beyond Scaffold Splits: Structural-Frontier Evaluation Reveals Hidden Failures in ADMET Models}
\date{*Email: \texttt{\url{karcenzheng@gmail.com}}}

\begin{document}

\maketitle

\begin{abstract}
\noindent
Molecular property models are routinely evaluated by holding out Bemis--Murcko
scaffolds, but a scaffold label captures only one notion of chemical novelty.  We
introduce a label-free structural-frontier split that reserves the sparsest and
most physicochemically remote scaffold groups, and benchmark it on six public
ADMET tasks against a ratio- and group-matched scaffold control.  The frontier
inflates primary error by a taskwise median of 87.0\% (skew-sensitive mean
130.3\%); the gap survives a message-passing graph network and, on average,
exceeds the published Lo-Hi and DataSAIL splits.  For blood--brain-barrier
permeability the perceptron head produces a genuine score-ranking inversion
rather than a prevalence artifact.  A count-adjusted multi-view tail-risk penalty and three fixed
distributionally robust objectives do not reliably close the gap.  Frontier
robustness therefore requires evidence beyond scaffold generalization, and split
construction and label provenance are first-order evaluation constraints.
\end{abstract}

\section*{Keywords}

ADMET; Cheminformatics; QSAR; Machine Learning; Applicability Domain

\section*{Introduction}

Machine-learning models for absorption, distribution, metabolism, excretion and
toxicity (ADMET) are used to prioritize compounds before expensive assays.
Prospective compounds, however, can occupy chemical regions that are sparsely
represented in historical data.  A random test partition therefore gives an
optimistic estimate when close analogues cross folds.  The widely used
Bemis--Murcko scaffold split mitigates this leakage by grouping molecules around a
common ring-and-linker framework \cite{bemis1996murcko}, but it converts a
continuous and representation-dependent support question into a single binary
partition.

Recent evaluations show that molecular ranking can change substantially with the
dataset, representation and split \cite{vanTilborg2023systematic}; simulated
prospective splits \cite{sheridan2023simpd}, low-similarity benchmarks
\cite{stanley2023lohi}, distance-aware evaluation \cite{realworldmood2024} and
leakage-controlled splitting \cite{datasail2025} each expose failure modes that an
ordinary scaffold holdout may miss.  That raises a sharper question: does a model
remain reliable at a \emph{structural frontier} that is simultaneously remote and
locally unsupported, and does that reliability survive different chemically
defensible representations?
Recent systematic splitter audits and property-tail benchmarks reach a broader
conclusion: no split or molecular model is uniformly best
\cite{fooladi2025molecularood,antoniuk2025boom}, while unfamiliarity itself predicts
performance near the edge of chemical space \cite{vantilborg2026edge}.

Label provenance deserves the same scrutiny.
\cmnpd, a database of marine natural products \cite{lyu2021cmnpd}, contains six
ADMET fields generated by BIOVIA Pipeline Pilot models, not by experiments
\cite{cmnpdParameters}.
Treating these fields as biological ground truth would turn evaluation into
distillation of a legacy teacher.  It would also make a claim of concept drift
unidentifiable: if the observed target is $\widetilde Y=h(X)$, a change in
$P(X)$ or in the marginal $P(\widetilde Y)$ does not establish a change in the
conditional mechanism.

We therefore separate two questions.  Biological evaluation uses six public,
measured or curated endpoint tasks distributed through Therapeutics Data Commons (TDC)
\cite{huang2021tdc}.  A secondary \cmnpd\ analysis measures only agreement with the
legacy teacher as natural-product novelty increases.  The study makes three
contributions.  First, it specifies a label-free 70/10/20 frontier split using
scaffold-level robust descriptor radius and local sparsity.  Its central control
uses the same 70/10/20 ratio and the same deterministic acyclic group units, but
allocates them by seeded size-aware scaffold splitting; a conventional 80/10/10
split is retained as context.  Second, it formulates \method\ as a falsifiable
probe that penalizes a count-adjusted tail and outward risk slope across ECFP,
MACCS, atom-pair and physicochemical views.  Third, it performs 468 fully recorded
primary runs with five paired seeds, fixed robust-penalty controls, ablations,
calibration metrics and linear controls, together with a 46-run message-passing
graph-network control that varies model capacity upward, plus an empirical
comparison against the Lo-Hi and DataSAIL splitters.  The matched frontier gap is
large, it survives a higher-capacity encoder, and it exceeds on average what two
published hard splits expose, while no tested penalty reliably closes it.  The
largest single-endpoint effect, BBB, is a genuine score-ranking inversion under
the perceptron head: it comes from the model's ordering of frontier molecules
rather than from the label-prevalence shift the frontier induces, and it
disappears under the graph network.  We therefore report the headline mean
alongside its skew-robust median and a leave-one-endpoint-out sensitivity.

\section*{Related work}

\paragraph{Molecular splits and chemical support.}
Scaffold splitting \cite{bemis1996murcko} is a useful leakage control, not a
complete deployment model.  SIMPD approximates medicinal-chemistry series
progression without timestamps \cite{sheridan2023simpd}; Lo-Hi constructs
low-similarity tests \cite{stanley2023lohi}; and Real-World MOOD relates
distance-to-training to predictive performance and calibration
\cite{realworldmood2024}.  DataSAIL optimizes leakage-aware partitions across
multiple domains \cite{datasail2025}.  Natural-product rankings also change across
fingerprint families \cite{boldini2024fingerprints}.  Our split is a
deterministic, label-free stress test of sparse
physicochemical support, reported beside the
standard scaffold benchmark.  We compare it empirically against Lo-Hi and DataSAIL
under a matched control in the \hyperref[sec:crosssplit]{cross-split comparison}.

\paragraph{Applicability domains and predictive uncertainty.}
Classical QSAR practice defines an applicability domain relative to a model,
representation and training set; descriptor-space definitions and their cutoffs
need not agree \cite{netzeva2005applicability,sahigara2012applicability,
sheridan2012domain}.  The robust radius and local-density ingredients in our score
inherit this distance-based AD lineage.  Our contribution is to use them as a
deterministic, label-free, ratio- and group-matched \emph{allocation stress test}
across molecular views, which yields an evaluation split rather than an
applicability-domain declaration or a per-compound confidence score.  Molecular
UQ benchmarks likewise find no estimator uniformly reliable across tasks
\cite{hirschfeld2020uq}.  Conformal prediction can provide finite-sample
marginal coverage under exchangeability \cite{angelopoulos2023conformal} and
has been proposed in QSAR as a transparent alternative to heuristic domain
cutoffs \cite{norinder2014conformal}; its ordinary guarantee does not survive the
deliberately shifted frontier, because exchangeability is exactly what the split
breaks.

\paragraph{Invariant and distributionally robust learning.}
Invariant risk minimization \cite{arjovsky2019irm}, GroupDRO
\cite{sagawa2020groupdro} and risk extrapolation \cite{krueger2021rex} protect
specified environment families; conditional value-at-risk (CVaR) emphasizes an
upper loss tail \cite{rockafellar2000cvar}.  Molecular methods have adapted
invariance objectives to graphs \cite{li2023imold}.  Efficient empirical GDRO
methods address structured observed groups \cite{yu2024minimax}, while recent
theory clarifies that robust risk may be only partially identified when deployment
shifts are not represented by training environments
\cite{partial2024robustness}.
Documented failures of graph invariance further caution against treating an
environment penalty as universal protection \cite{gui2024invariance}.  \method\
combines familiar ingredients to test a specific ambiguity family.

\paragraph{ADMET data provenance.}
TDC provides standardized therapeutic tasks and dataset metadata
\cite{huang2021tdc,tdc2020data}.  \cmnpd\ catalogues marine natural products
\cite{lyu2021cmnpd}, but its ADMET columns are computational annotations from
BIOVIA Pipeline Pilot 18.1 \cite{cmnpdParameters}.  Modern
systems such as ADMET-AI \cite{swanson2024admetai} and ADMETlab 3.0
\cite{fu2024admetlab} remain predictors rather than experimental oracles.  We
therefore reserve the word \emph{accuracy} for public endpoint labels and use
\emph{teacher agreement} for \cmnpd; the later audit bridges only explicitly
documented, approximate endpoint mappings.

\section*{Results}

\begin{figure}[t]
  \centering
  \includegraphics[width=\linewidth]{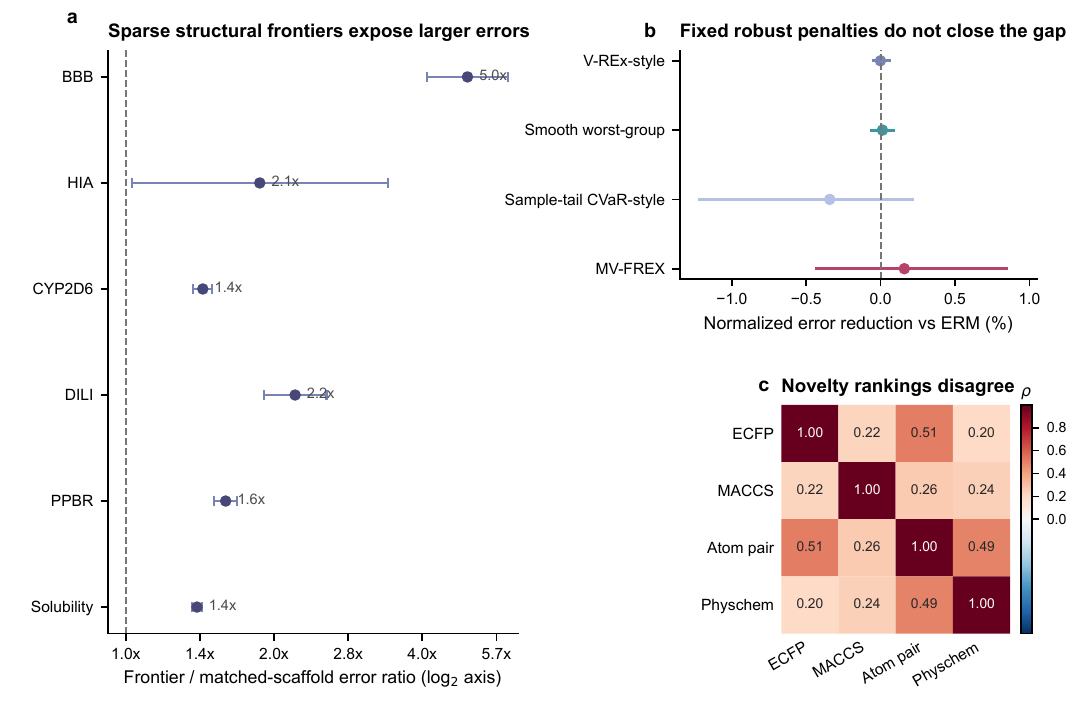}
  \caption{\textbf{Structural-frontier allocation dominates objective choice.}
  \textbf{a}, Frontier-to-matched-scaffold primary-error ratios for ERM; classification
  error is $1-$AUROC or $1-$AUPRC and regression error is MAE. Points are five-seed
  means and bars are standard deviations. \textbf{b}, Descriptive hierarchical-bootstrap mean
  normalized error reduction relative to frontier ERM (95\% stability intervals).
  \textbf{c}, Mean Spearman correlations among novelty views across tasks.  The
  split effect is large, whereas every fixed-penalty interval contains zero.}
  \label{fig:main}
\end{figure}

\subsection{The frontier exposes a large, endpoint-wide gap}

Every endpoint deteriorates under the label-free frontier split
(Table~\ref{tab:primary}; Fig.~\ref{fig:main}a).  Across equally weighted
endpoints the taskwise \emph{median} relative error inflation over the ratio- and
group-matched scaffold control is 87.0\%; the mean is 130.3\% (95\%
task/seed stability interval, 52.1--246.0\%; 30 task--seed effects) but is
right-skewed and dominated by one endpoint.  A leave-one-endpoint-out check
(Fig.~\ref{fig:bbb}a; Supporting Information, Table~S2) confirms
this: removing BBB drops the mean to 75.9\% and the median to 59.7\%, whereas
removing any other endpoint leaves the mean between 131.8\% and 148.4\%.  The
median and the BBB-excluded mean therefore summarize the effect, while the
six-endpoint mean is dominated by BBB.
The effect ranges from BBB AUROC falling from $0.879\pm0.020$ to
$0.409\pm0.005$ (a ranking inversion analysed in the
\hyperref[sec:bbb]{BBB sensitivity analysis}) to solubility
MAE rising from $1.373\pm0.040$ to $1.915\pm0.022$.  Chance-adjusting CYP2D6
AUPRC reduces its taskwise inflation from 43.5\% to 31.3\%, but the equally
weighted mean remains 128.2\% (interval, 48.7--244.7\%).  CYP2D6 AUROC error rises
by 42.6\%, and a three-seed linear control shows a mean 112.4\% gap (interval,
36.7--229.1\%).

\begin{table}[t]
  \caption{\textbf{Primary performance under scaffold and frontier allocations.}
  The matched scaffold column uses the frontier's 70/10/20 ratio and acyclic group
  units; the conventional 80/10/10 result is contextual. Values are mean $\pm$
  standard deviation over five seeds.}
  \label{tab:primary}
  \centering
  \small
  \resizebox{\linewidth}{!}{\begin{tabular}{llcccc}
\toprule
Task & Metric & Scaffold 80/10/10 & Matched scaffold & Frontier ERM & Frontier MV-FREX \\
\midrule
BBB & AUROC $\uparrow$ & 0.885 $\pm$ 0.019 & 0.879 $\pm$ 0.020 & 0.409 $\pm$ 0.005 & 0.409 $\pm$ 0.004 \\
HIA & AUROC $\uparrow$ & 0.692 $\pm$ 0.169 & 0.717 $\pm$ 0.161 & 0.538 $\pm$ 0.015 & 0.535 $\pm$ 0.024 \\
CYP2D6 & AUPRC $\uparrow$ & 0.581 $\pm$ 0.027 & 0.569 $\pm$ 0.017 & 0.383 $\pm$ 0.005 & 0.383 $\pm$ 0.005 \\
DILI & AUROC $\uparrow$ & 0.747 $\pm$ 0.046 & 0.811 $\pm$ 0.025 & 0.584 $\pm$ 0.029 & 0.589 $\pm$ 0.033 \\
PPBR & MAE $\downarrow$ & 10.457 $\pm$ 1.349 & 11.234 $\pm$ 0.653 & 17.897 $\pm$ 0.104 & 17.869 $\pm$ 0.100 \\
Solubility & MAE $\downarrow$ & 1.364 $\pm$ 0.028 & 1.373 $\pm$ 0.040 & 1.915 $\pm$ 0.022 & 1.917 $\pm$ 0.015 \\
\bottomrule
\end{tabular}
}
\end{table}

The split is label-free, and structural selection still shifts the label
distribution.  In seed 0, positive
prevalence changes from train to frontier test from 86.0\% to 48.1\% for BBB, 97.0\%
to 52.6\% for HIA, 20.9\% to 13.0\% for CYP2D6 and 53.0\% to 40.7\% for DILI.
These shifts follow from how molecules are allocated and do not by themselves
establish conditional or causal concept drift.

\subsection{The BBB effect is a ranking inversion, not a prevalence artifact}
\label{sec:bbb}

Because BBB drives the mean, we examine it directly.  Its frontier-test AUROC is
$0.409\pm0.005$ over five seeds sharing an identical 389-molecule test set (187
positive, 202 negative), with a compound-and-seed bootstrap interval of
0.352--0.467 in which only 0.155\% of resamples reach chance.  AUROC is a rank
statistic and is invariant to class prevalence, so this sub-chance value is a
genuine \emph{ranking inversion}: the model orders frontier molecules opposite to
their labels, and the prevalence shift above cannot arithmetically produce it.
Exact class-prevalence reweighting confirms this: the standardized AUROC is
unchanged at 0.409, while the prevalence-sensitive metrics move as expected
(Supporting Information, Table~S4).

\begin{figure}[t]
  \centering
  \includegraphics[width=\linewidth]{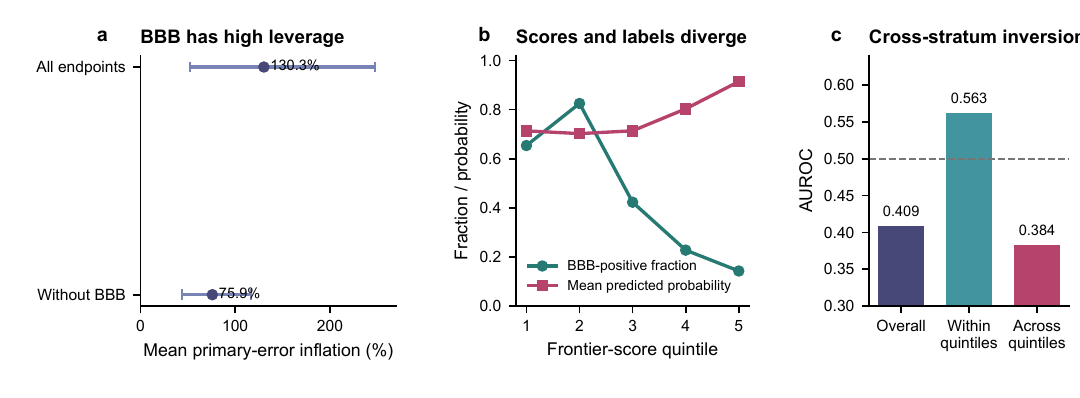}
  \caption{\textbf{BBB drives the headline mean, and its sub-chance AUROC is a rank
  inversion rather than a prevalence effect.} \textbf{a}, Leaving BBB out lowers the
  mean primary-error inflation from 130.3\% to 75.9\% (task/seed
  intervals). \textbf{b}, Across frontier-score quintiles the BBB-positive fraction
  falls while the mean predicted probability rises, so scores and labels move in
  opposite directions. \textbf{c}, Ordering stays above chance within a novelty
  stratum but inverts across strata, which yields the sub-chance overall AUROC.}
  \label{fig:bbb}
\end{figure}

The mechanism is a support-confounded association.  The true BBB label correlates
negatively with the frontier score (Spearman $-0.455$) while the model's
prediction correlates positively (Spearman $+0.369$): as molecules become sparser
and more remote they become less permeable, but the model scores them as more
permeable (Fig.~\ref{fig:bbb}b; Supporting Information, Table~S3:
mean prediction rises from 0.71 to 0.91 as the positive fraction falls from 0.65 to
0.14).
Ordering stays above chance \emph{within} a novelty stratum (within-bin AUROC
0.563) but inverts \emph{across} strata (cross-bin AUROC 0.384; Fig.~\ref{fig:bbb}c);
frontier \emph{validation} AUROC is 0.805, so early stopping was not the cause.
The failure is one of extrapolated support, and it is the
perceptron head's most extreme behaviour rather than a typical one.

\subsection{A higher-capacity encoder reproduces the gap without the inversion}
\label{sec:gnn}

\begin{figure}[t]
  \centering
  \includegraphics[width=\linewidth]{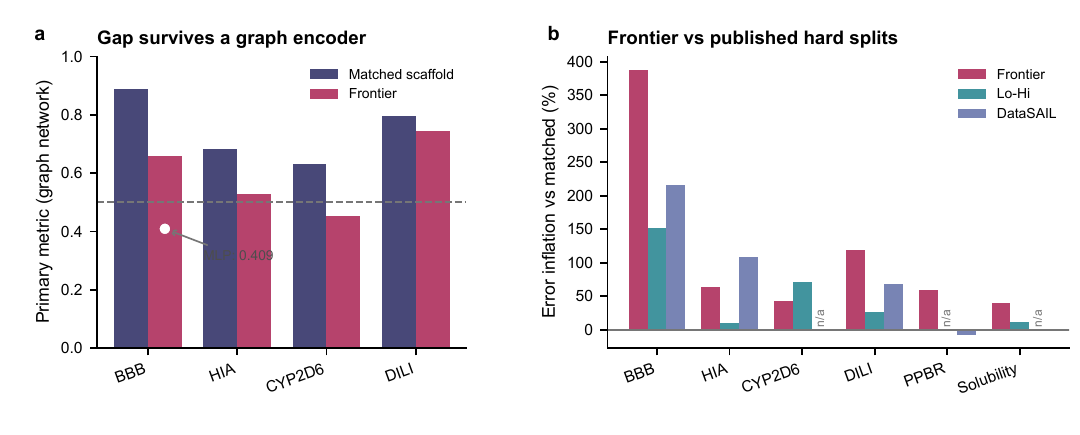}
  \caption{\textbf{The matched frontier gap is robust to model capacity and exceeds
  two published hard splits.} \textbf{a}, Under a message-passing graph network the
  frontier still degrades every task relative to the matched scaffold control; the
  open marker shows that the perceptron drove BBB to 0.409, whereas the graph
  network keeps it above chance at 0.659. \textbf{b}, Primary-error inflation over
  the matched control for the frontier, Lo-Hi and DataSAIL. The frontier is the
  most severe on average, but the splits are not interchangeable: DataSAIL makes HIA
  harder and PPBR easier, while Lo-Hi is infeasible for PPBR (marked n/a, as are the
  DataSAIL tasks left unsolved within budget).}
  \label{fig:robust}
\end{figure}

Replacing the fixed-fingerprint perceptron with a message-passing graph network
(304,260 parameters) on the four binary tasks leaves the matched frontier gap
large: the mean normalized error inflation is 82.8\% (task/seed interval,
21.8--187.2\%; four tasks, 15 paired effects), and every task still degrades from
the matched scaffold control (Table~\ref{tab:gnn}; Fig.~\ref{fig:robust}a).  A
low-capacity head is therefore not the source of the effect.  Two things change
with capacity.  First,
BBB no longer inverts: its frontier AUROC rises from 0.409 to $0.659\pm0.093$,
confirming that the sub-chance ranking was specific to the weakest model and that
the endpoint-wide gap does not depend on it.  Second, \method\ remains
inconclusive under the stronger encoder, changing normalized frontier error by
$-1.9\%$ (interval, $-8.0$--3.9\%); the penalty null therefore holds across two
model families.

\begin{table}[t]
  \caption{\textbf{Graph-network capacity control.}  Primary metric under the
  matched scaffold control and the frontier split; values are mean $\pm$ standard
  deviation over five seeds.  $^\dagger$DILI reports a single completed seed and is
  descriptive.}
  \label{tab:gnn}
  \centering
  \small
  \begin{tabular}{llccc}
\toprule
Task & Metric & Matched scaffold & Frontier ERM & Frontier MV-FREX \\
\midrule
BBB & AUROC $\uparrow$ & 0.888 $\pm$ 0.034 & 0.659 $\pm$ 0.093 & 0.680 $\pm$ 0.052 \\
HIA & AUROC $\uparrow$ & 0.683 $\pm$ 0.028 & 0.527 $\pm$ 0.028 & 0.527 $\pm$ 0.028 \\
CYP2D6 & AUPRC $\uparrow$ & 0.632 $\pm$ 0.011 & 0.461 $\pm$ 0.029 & 0.456 $\pm$ 0.031 \\
DILI$^\dagger$ & AUROC $\uparrow$ & 0.765 & 0.733 & 0.704 \\
\bottomrule
\end{tabular}

\end{table}

\subsection{The frontier exposes larger and distinct failures than existing hard splits}
\label{sec:crosssplit}

To place the frontier among established difficult splits, we ran ERM on the same
harmonized molecule sets under two published low-similarity splitters, Lo-Hi
\cite{stanley2023lohi} and DataSAIL \cite{datasail2025}, using their official
implementations, and measured primary-error inflation over the identical matched
scaffold control (Table~\ref{tab:crosssplit}).  DataSAIL reproduces the 70/10/20
ratio closely (realized test fraction 0.19).  Lo-Hi discards near-neighbour
molecules to enforce dissimilarity and returns a smaller test block (0.14); its
mixed-integer program was infeasible for PPBR at any feasible fraction, so that
cell is empty.  Two conclusions follow.

First, the frontier is on average the most severe (Fig.~\ref{fig:robust}b): mean
error inflation over the matched control is 118.9\% for the frontier versus 96.0\%
for DataSAIL and 54.1\% for Lo-Hi (each over that splitter's covered tasks;
restricting the frontier to the same tasks preserves the ordering, at 157.6\% on
DataSAIL's four and 130.8\% on Lo-Hi's five).  Only the frontier drives BBB below chance (AUROC 0.409, against
0.695 for Lo-Hi and 0.618 for DataSAIL), so the
\hyperref[sec:bbb]{BBB inversion} is specific to sparse
physicochemical support rather than a
generic consequence of any hard split.  Second, the splitters are not
interchangeable: DataSAIL makes HIA harder than the frontier does (AUROC 0.412
versus 0.538) yet leaves PPBR slightly easier than the matched control
($-7.6\%$), and Lo-Hi makes CYP2D6 harder than the frontier ($+71\%$ versus
$+43\%$).  No single split is uniformly hardest, so we recommend
reporting a support frontier \emph{beside} existing splits.

\begin{table}[t]
  \caption{\textbf{Frontier versus published hard splits (ERM).}  Primary metric
  under each split, mean over five (frontier, matched) or three (Lo-Hi, DataSAIL)
  seeds; bottom rows give mean and median error inflation over the matched scaffold
  control across covered tasks.  Lo-Hi was infeasible for PPBR; DataSAIL was run on
  the four tasks whose mixed-integer split solved within budget.  Realized ratios:
  DataSAIL $\approx$70/10/20, Lo-Hi $\approx$75/10/14.}
  \label{tab:crosssplit}
  \centering
  \small
  \begin{tabular}{llcccc}
\toprule
Task & Metric & Matched scaffold & Frontier & Lo-Hi & DataSAIL \\
\midrule
BBB        & AUROC $\uparrow$ & 0.879 & \textbf{0.409} & 0.695 & 0.618 \\
HIA        & AUROC $\uparrow$ & 0.717 & 0.538 & 0.691 & \textbf{0.412} \\
CYP2D6     & AUPRC $\uparrow$ & 0.569 & 0.383 & 0.262 & --- \\
DILI       & AUROC $\uparrow$ & 0.811 & 0.584 & 0.760 & 0.682 \\
PPBR       & MAE $\downarrow$ & 11.234 & 17.897 & --- & 10.385 \\
Solubility & MAE $\downarrow$ & 1.373 & 1.915 & 1.537 & --- \\
\midrule
\multicolumn{2}{l}{Mean error inflation vs.\ matched (\%)} & --- & 118.9 & 54.1 & 96.0 \\
\multicolumn{2}{l}{Median error inflation vs.\ matched (\%)} & --- & 61.4 & 26.5 & 87.9 \\
\bottomrule
\end{tabular}

\end{table}

\subsection{Fixed robust penalties do not reliably close the gap}

\method\ reduces normalized frontier error by 0.160\% versus ERM (95\%
stability interval, $-0.429$--0.844\%), with 16 wins and 14 losses across 30 paired
comparisons (Fig.~\ref{fig:main}b).  Its worst-view-bin reduction is 1.30\%
(interval, $-0.89$--4.89\%).  The V-REx-style, smooth worst-group and sample-tail
CVaR-style penalties are also mixed (Table~\ref{tab:methods}); every interval
includes zero.  Removing the
ordered slope is slightly better for DILI and nearly tied elsewhere, while the
full objective is marginally better for PPBR.  A linear head and the graph network
of the \hyperref[sec:gnn]{graph-network control} both show the same flat objective
effect.  The motivating hypothesis was that \method\ would improve frontier risk
broadly, and the controls leave frontier error essentially unchanged.  Penalty
hyperparameters are fixed; tuning them against the frontier would select on the
evaluation target.

\begin{table}[t]
  \caption{\textbf{Frontier objective effects relative to ERM.}
  Positive values indicate lower normalized primary error; intervals use the
  descriptive hierarchical task/seed bootstrap.}
  \label{tab:methods}
  \centering
  \small
  \begin{tabular}{lrrr}
\toprule
Method & Error reduction (\%) & 95\% CI (\%) & Win / tie / loss \\
\midrule
V-REx-style & -0.001 & [-0.049, 0.061] & 12 / 5 / 13 \\
Smooth worst-group & 0.012 & [-0.060, 0.089] & 16 / 1 / 13 \\
Sample-tail CVaR-style & -0.342 & [-1.214, 0.215] & 17 / 0 / 13 \\
MV-FREX & 0.160 & [-0.429, 0.844] & 16 / 0 / 14 \\
\bottomrule
\end{tabular}

\end{table}

\subsection{Chemical unfamiliarity depends on the molecular view}

Pairwise Spearman correlations between frontier-test novelty scores range from
$-0.093$ to 0.837 (mean 0.321); ECFP--physicochemical correlation averages 0.198
(Fig.~\ref{fig:main}c).  The frontier score's strongest correlation with any one
descriptor is only 0.31--0.55 across tasks.  This level of disagreement between
views is why the frontier is evaluated under several representations, and it
limits what any single-view risk penalty can be expected to cover.

\begin{figure}[t]
  \centering
  \includegraphics[width=\linewidth]{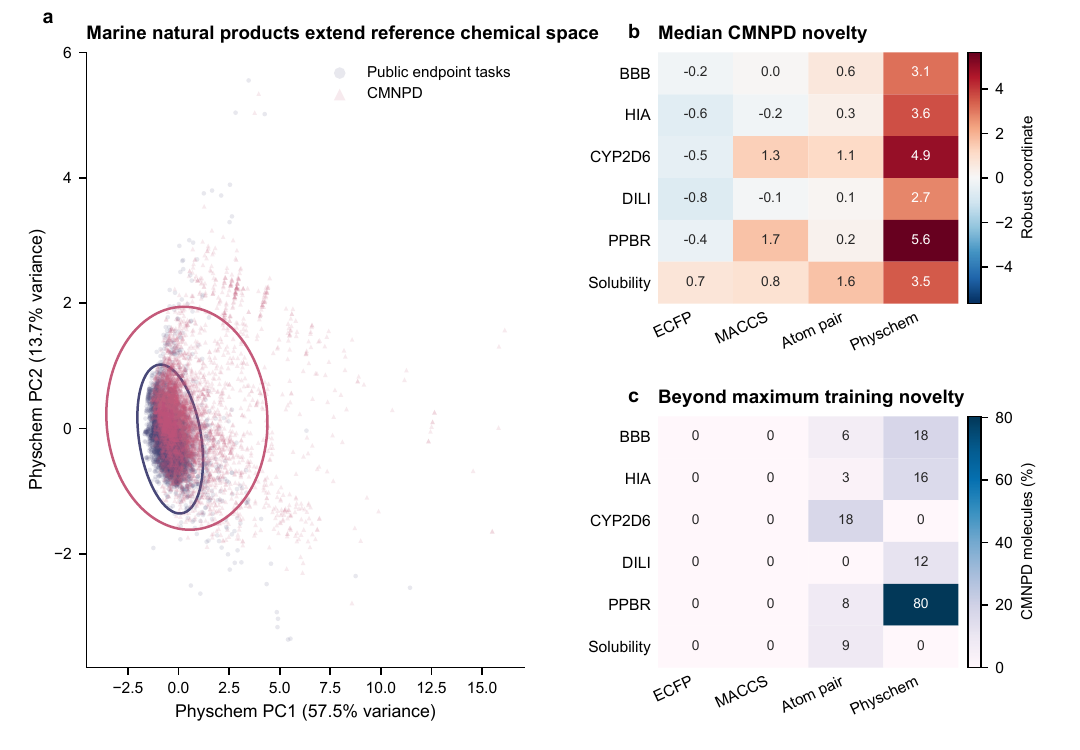}
  \caption{\textbf{Marine natural products are representation-dependent OOD
  targets.} \textbf{a}, Descriptive PCA of robust-scaled physicochemical
  descriptors for deterministic samples from the six public endpoint tasks and
  \cmnpd; outlines are 95\% covariance ellipses. \textbf{b}, Median \cmnpd\
  novelty coordinate by task and view. \textbf{c}, Fraction of \cmnpd\ beyond the
  maximum task-training coordinate.  PCA is illustrative; panels b--c carry the
  quantitative view-specific comparison.}
  \label{fig:cmnpdshift}
\end{figure}

\section*{\cmnpd\ teacher-agreement audit}

All 31,561 \cmnpd\ SMILES parse, but its six ADMET fields are Pipeline Pilot
predictions \cite{lyu2021cmnpd,cmnpdParameters}.  We apply five-seed ensembles trained on the public endpoint tasks
and score only legacy-teacher agreement.  The mappings are approximate: the BBB
and HIA teachers use level thresholds, the CYP field is a legacy binary model
trained separately from the TDC inhibition labels, PPBR is a binary teacher compared with a thresholded
regression predictor, and hepatotoxicity is not identical to curated human DILI.
The Supporting Information records every mapping and missing-value rule
(Table~S8) and summarizes overall ensemble agreement
(Table~S9).

\cmnpd\ can appear close in ECFP space yet remote in physicochemical space
(Fig.~\ref{fig:cmnpdshift}).  Its task-referenced median physicochemical novelty
coordinates range from 2.73 to 5.63, and 80.3\% of \cmnpd\ exceeds the maximum
PPBR training coordinate in that view.  Teacher availability itself changes with
novelty (Supporting Information, Fig.~S2): the usable BBB fraction falls from 97.4\% in
quintile Q1 to 0.73\% in Q5.
BBB teacher-agreement AUROC decreases from 0.592 (compound-bootstrap interval,
0.577--0.607) to 0.332 (0.156--0.528), but the latter uses only 46 compounds
(10 positive, 36 negative) and is descriptive under this coverage collapse.
For solubility, full-range teacher MAE rises from 1.160 to 5.434; the descriptive
$|\widetilde y|\leq10$ analysis rises from 1.160 to 3.349 while retaining 93.1\%
of Q5, and Spearman correlation falls from 0.609 to $-0.012$.  CYP2D6 is a
counterexample whose agreement improves with novelty.  Neither direction
identifies biological accuracy.

\section*{Discussion}

We set out to ask whether an ADMET model that passes a scaffold holdout stays
reliable at a \emph{structural frontier} that is at once remote and locally
unsupported, and whether that answer survives a change of chemically defensible
representation.  Both answers are negative.  Ranking otherwise identical
scaffold-group units by sparse physicochemical support inflates equally weighted
primary error by a taskwise median of 87.0\% over the ratio- and group-matched
control (skew-sensitive mean 130.3\%), and the gap persists under a linear head and
a message-passing graph network (mean 82.8\% across four tasks).  Reliability is
representation-dependent as well: pairwise Spearman correlations between
view-specific novelty scores average only 0.321, and ECFP and physicochemical
novelty agree at 0.198.  A model can look safe on a scaffold benchmark, and even on
one notion of chemical distance, yet fail on another.  This turns the emerging
consensus that no split or model is uniformly best \cite{fooladi2025molecularood,
antoniuk2025boom} into something more specific: the failure is largest exactly
where local support is sparse, the regime \cite{vantilborg2026edge} identify at
the edge of chemical space.

\subsection*{Why the tested robust objectives do not help}
The flat result on training penalties is the most informative outcome here.
\method\ changes normalized frontier error
by 0.16\% under the perceptron and by $-1.9\%$ under the graph network, and the
V-REx-style, smooth worst-group and CVaR-style controls are equally inconclusive.
Recent theory anticipates this.  \cite{partial2024robustness} show that robust
risk is only partially identified once the deployment shift is not represented by
the training environments, and our environments are quantile bins of
\emph{source-view} novelty while the frontier is, by construction, a target that
lies outside those bins.  Proposition~\ref{prop:control} states the same boundary
operationally: its guarantee holds for a declared mixture of observed view--bin
conditionals, and the frontier is not such a mixture, so an outward risk slope
fitted on source bins has no reason to extrapolate.  The ablation bears this out,
since removing the slope barely moves the result.  Documented failures of graph
invariance under shift \cite{gui2024invariance} and the broader difficulty of
learning invariant molecular representations \cite{li2023imold,yu2024minimax}
point the same way: an environment penalty protects the environments it can see, not
unseen chemistry.  Frontier robustness is therefore a
data-support problem before it is an objective-design problem.

\subsection*{Placing the structural frontier among existing OOD splits}
Our matched comparison locates the frontier among published difficult splits
without confounding difficulty with test size or grouping granularity.  Averaged
over the tasks each method covers, the frontier inflates error more than DataSAIL
\cite{datasail2025} or Lo-Hi \cite{stanley2023lohi} (118.9\% against 96.0\% and
54.1\%), and it alone drives an endpoint below chance.  The splitters are not
interchangeable, though: DataSAIL makes HIA harder and PPBR easier, and Lo-Hi could
not form a low-similarity PPBR partition at all.  Each stresses a different axis of
chemical space, which is why we report the frontier beside the scaffold benchmark
and complementary designs such as the simulated time splits of \cite{sheridan2023simpd} and the distance-aware evaluation of
\cite{realworldmood2024}.  What the frontier adds is
methodological control: because it reuses the matched split's ratio and acyclic
group units, the excess error it exposes is attributable to \emph{where} molecules
are held out, not to how many.

The frontier is, in this sense, a modern applicability domain.  Its robust-radius and
local-density ingredients descend from distance-based AD definitions
\cite{netzeva2005applicability,sahigara2012applicability,sheridan2012domain}, and
the BBB inversion shows what an AD violation looks like once it is scored rather than
flagged: within a novelty stratum the model still ranks above chance (AUROC 0.563),
but across strata the learned association reverses (0.384), so the pooled ranking
falls to 0.409.  Conformal prediction is often offered as a transparent alternative
to hard AD cutoffs \cite{norinder2014conformal,angelopoulos2023conformal}, but its
marginal coverage assumes exchangeability, which the frontier deliberately breaks;
together with the finding that molecular uncertainty estimators are not uniformly
reliable \cite{hirschfeld2020uq}, calibrated abstention at the frontier remains an
open problem.

\subsection*{Measured versus predicted labels}
The \cmnpd\ audit measures teacher agreement.  As set out in the
Introduction, if the observed target is $\widetilde Y=h(X)$ then a change in $P(X)$
or in $P(\widetilde Y)$ cannot identify a change in the biological conditional, so
treating the BIOVIA Pipeline Pilot fields as ground truth would only measure how
well a new model distils a legacy one.  The audit still carries information:
agreement falls with novelty for BBB and solubility, improves for CYP2D6, and BBB
teacher coverage collapses from 97.4\% in the nearest quintile to 0.73\% in the
farthest.  Each of these describes two predictors disagreeing, not a biological
outcome, which is why we reserve the word
\emph{accuracy} for the measured TDC endpoints \cite{huang2021tdc} and use
\emph{teacher agreement} for the predicted natural-product fields.

\subsection*{Limitations}
Five boundaries constrain the conclusions.  The frontier is a controlled
descriptor-space stress test, not a timestamped prospective campaign, and its outer
tail shifts label prevalence.  The descriptive task/seed bootstrap resamples only
six endpoint units and captures algorithmic stability rather than laboratory
sampling error, so the six-endpoint mean should always be read next to its median
and leave-one-endpoint-out range.  The capacity control spans four binary tasks and
the ERM/\method\ contrast; regression endpoints, large pretrained encoders and the
three remaining penalties remain untested under a graph network.  The
external comparison inherits each splitter's native behaviour, including Lo-Hi's
smaller realized test fraction and its PPBR infeasibility.  Finally, the \cmnpd\
fields are computational, its endpoint mappings are approximate, and its coverage
collapses at the frontier, so that analysis cannot speak to biological accuracy.

\subsection*{Recommendations and outlook}
Two low-cost practices would make molecular OOD evaluation more informative.
Benchmark maintainers such as TDC \cite{huang2021tdc} could pair every scaffold split with at
least one label-free support frontier, release split manifests and label
distributions, and report performance under more than one novelty view, since a
single fingerprint family can hide a failure that another exposes
\cite{boldini2024fingerprints,vanTilborg2023systematic}.  Method developers, in
turn, should treat a robust objective as a claim about named environments and test
it against a target shift it has never seen instead of assuming that it transfers.
The most valuable next step is prospective: assaying newly synthesised natural
products would let representation uncertainty, assay shift and biological accuracy
be evaluated together, converting the frontier from a stress test into a measured
forecast of deployment risk.
\section*{Conclusion}

Across six public measured or curated ADMET tasks, a structural-frontier holdout
increases equally weighted primary error by a median of 87.0\% (skew-sensitive
mean 130.3\%) relative to a ratio- and group-matched scaffold control.  This gap
survives a higher-capacity graph-network encoder, while \method\ and three fixed
robust-penalty implementations leave it unchanged under every model we
tried.  A separate natural-product audit shows that both OOD status and
legacy-teacher agreement vary with representation and endpoint.  Frontier
robustness therefore has to be demonstrated separately from scaffold
generalization, and experimental validation remains the reference for database
annotations.

\section*{Computational Methods}

\subsection*{Matched label-free benchmarks}

For task $t$, let $D_t=\{(x_i,y_i)\}_{i=1}^{n_t}$ contain a molecular graph and a
public experimental or curated binary or continuous endpoint.  Connectivity-equivalent structures are
harmonized before splitting; binary replicates use majority vote with exact ties
removed, and regression replicates use the median.  The conventional reference
assigns stereochemistry-free Murcko groups in an 80/10/10 ratio and retains the
literal empty scaffold as one disclosed acyclic group.

The central comparison instead uses identical grouping units and 70/10/20 ratios.
RDKit descriptors are robustly scaled, and only the heterogeneous empty-scaffold
group is partitioned into deterministic descriptor-space cells of at most 50
molecules.  The matched scaffold control allocates these units by seeded,
size-aware scaffold splitting.  The frontier allocation ranks the same units.  For
scaffold $s$, let $c_s$ be the median descriptor vector, $r_s$ its distance from
the median scaffold and $d_s$ its mean distance to the ten nearest scaffold
centres.  With $\operatorname{rank}_{01}$ denoting a rank normalized to $[0,1]$,
\begin{equation}
q_s=\tfrac12\operatorname{rank}_{01}(r_s)
    +\tfrac12\operatorname{rank}_{01}(d_s).
\label{eq:frontier}
\end{equation}
The highest-scoring groups supply approximately 20\% test molecules, the next 10\%
validation and the remaining 70\% training.  Thus the matched contrast changes
allocation by $q_s$, not sample ratio or acyclic grouping.  Labels are not read by
either procedure.

\subsection*{Multi-view ordered environments}

All predictors receive the same 1,024-bit radius-2 Morgan fingerprint
\cite{rogers2010ecfp}.  Additional views
\begin{equation}
\mathcal V=\{\mathrm{ECFP},\mathrm{MACCS},\mathrm{atom\ pair},
\mathrm{physchem}\} 
\end{equation}
define risk environments only.  Fingerprints are reduced to
at most 32 training-fitted singular-vector components; physicochemical descriptors
are robustly scaled.  For view $v$, training median $m_v$ and mean five-neighbour
distance $d_{v,5}$, the novelty coordinate is
\begin{equation}
z_v(x)=\tfrac12\,\operatorname{rscale}
  \!\left(\|\psi_v(x)-m_v\|_2\right)
+\tfrac12\,\operatorname{rscale}\!\left(d_{v,5}(x)\right),
\label{eq:novelty}
\end{equation}
where each median and median absolute deviation is fitted on training molecules.
Training quartiles create ordered bins $b\in\{1,\ldots,4\}$; held-out molecules
are assigned with frozen thresholds.  The frontier allocation score
$q_s$ (Eq.~\ref{eq:frontier}) and the physicochemical view $z_{\mathrm{physchem}}$
are built from the same RDKit descriptor family, so this environment is partly
aligned with the axis that constructs the split; the penalty's remaining three
views (ECFP, MACCS, atom pair) are not, and the weak inter-view rank agreement
reported below quantifies how far apart they are.

\subsection*{A bounded, view-enveloped risk objective}

Binary tasks use cross-entropy; regression targets are standardized on training
data and use Huber loss.  Only the robust term is clipped:
$\bar\ell_i=\min\{\ell_i,5\}/5\in[0,1]$; empirical risk remains unclipped.  Let
$N_{v,b}$ be the full training count in one of $G=16$ prespecified view--bin
groups.  On minibatch $\mathcal M$, the executable count-adjusted quantity is
\begin{equation}
\widetilde U^{\mathcal M}_{v,b}=
\frac{1}{|\mathcal M_{v,b}|}\sum_{i\in\mathcal M_{v,b}}\bar\ell_i+
\sqrt{\frac{\log(2G/\delta)}{2N_{v,b}}},\qquad \delta=0.05.
\label{eq:batchrisk}
\end{equation}
Groups absent from a minibatch are omitted.  With
$k=\lceil0.4\times4\rceil$, $\widetilde T^{\mathcal M}$ is the maximum, over
views, of the mean of the $k$ largest finite
$\widetilde U^{\mathcal M}_{v,b}$ values.  $\widetilde S^{\mathcal M}$ is the
maximum positive secant slope between ordered bin centres.  The executable
\method\ objective is
\begin{equation}
\mathcal J_{\mathcal M}(\theta)=
\frac{1}{|\mathcal M|}\sum_{i\in\mathcal M}\ell_i(\theta)
+0.5\widetilde T^{\mathcal M}(\theta)
+0.2\widetilde S^{\mathcal M}(\theta).
\label{eq:objective}
\end{equation}
Finite maxima use a temperature-0.1 log-sum-exp.  Because minibatch means are
combined with full training counts, Eq.~\ref{eq:batchrisk} is a regularizer, not
an unbiased full-sample objective or post-selection confidence certificate.  It
changes no encoder or predictor capacity.

\begin{proposition}[Coverage for a declared mixture family]
\label{prop:control}
Fix a predictor, all view maps and all bin thresholds independently of an i.i.d.\
evaluation sample, and assume every declared view--bin group is represented.
Assume also that the ordered bin centres $c_{v,b}$ are distinct.  Let
$P_{v,b}=P(\,\cdot\mid B_v=b)$ and compute
$\widehat U^{E}_{v,b}=\widehat R^{E}_{v,b}+
\sqrt{\log(2G/\delta)/(2n^{E}_{v,b})}$ from bounded evaluation losses.  With
$k=\lceil0.4\times4\rceil$, define
\[
T^E=\max_v\frac1k\sum_{b\in\TopK_k(\widehat U^E_v)}\widehat U^E_{v,b},
\qquad
S^E=\max_{v,\,b<b'}
\left[\frac{\widehat U^E_{v,b'}-\widehat U^E_{v,b}}
{c_{v,b'}-c_{v,b}}\right]_+ .
\]
With probability at least $1-\delta$, for every declared view and every target
$Q_0=\sum_bq_bP_{v,b}$ satisfying $q_b\geq0$, $\sum_bq_b=1$ and $q_b\leq1/k$,
$R_{Q_0}\leq T^E$.  Moreover, for fixed $\rho,\eta\geq0$, define the declared
extension family to contain only targets $Q$ for which such a $Q_0$ exists and
$R_Q-R_{Q_0}\leq\rho(S^E+\eta)$.  Every member then satisfies
$R_Q\leq T^E+\rho(S^E+\eta)$.
\end{proposition}
\noindent The proof is given in the Supporting Information.  The condition
$Q_0=\sum_bq_bP_{v,b}$ excludes within-bin covariate or label-mechanism change,
and membership in the extension family is assumed rather than inferred from
source data.  The proposition is delimiting, not predictive: it
states precisely the declared family for which the evaluation-time quantities
$T^E,S^E$ are valid upper bounds, and the realized simultaneous Hoeffding
correction (0.035--0.202 across groups; Supporting Information) shows the
bound is non-vacuous but loose.  It leaves the training objective, unrestricted
molecular OOD and the frontier gap itself to the runs below.

\subsection*{Experimental design}

\paragraph{Tasks and metrics.}
We use public experimental or curated tasks for BBB permeability (1,941
molecules), intestinal absorption (577), CYP2D6 inhibition (12,888), human plasma
protein binding (1,578; percentage points), aqueous solubility (9,690;
$\log_{10}$ mol l$^{-1}$) and drug-induced liver injury (475).  Classification
uses AUROC except imbalanced CYP2D6, for which AUPRC is primary; regression uses
mean absolute error (MAE).  Secondary metrics include AUPRC, balanced accuracy,
Matthews correlation, Brier score and expected calibration error for
classification, and RMSE, Spearman correlation and $R^2$ for regression.
The standardized tables trace to the original BBB, HIA, CYP, PPBR, AqSolDB and
DILI sources \cite{martins2012bbb,hou2007hia,veith2009cyp,wenlock2015ppbr,
sorkun2019aqsoldb,xu2015dili,tdc2020data}.

\paragraph{Models and controls.}
The common predictor is a 64-unit ReLU multilayer perceptron with 0.1 dropout.
AdamW uses learning rate 0.002, weight decay $10^{-4}$, batches up to 1,024,
gradient clipping at 5, at most 30 epochs and validation patience 6.  We compare
empirical risk minimization (ERM), an ECFP V-REx-style variance penalty, an ECFP
smooth worst-group penalty, a minibatch sample-tail CVaR-style penalty and
\method\ under identical splits, batches and seeds.  These are fixed
implementations, not a hyperparameter search or canonical reproduction of every
published algorithm.  Three-seed ablations remove the slope or retain only the
ECFP view; a linear fingerprint head tests downward capacity dependence.  To test
upward capacity dependence, a separate control replaces the fingerprint head with
a message-passing graph neural network (304,260 parameters) that reads molecular
graphs directly, run under the same splits, seeds and ERM/\method\ objectives on
the four binary tasks with sufficient positive support (BBB, HIA, CYP2D6, DILI).
Five paired seeds jointly control split tie-breaking, initialization, dropout and
minibatch order.

\paragraph{Statistics and integrity.}
Tables report mean $\pm$ standard deviation over five complete runs.  For primary
score $m$, define error as $e=1-m$ for classification and $e=m$ for regression;
the matched split effect is $(e_{\mathrm{frontier}}-e_{\mathrm{scaffold}})/
e_{\mathrm{scaffold}}$.  For CYP2D6, a sensitivity analysis uses
$e=(1-\mathrm{AUPRC})/(1-p_{\mathrm{test}})$ and a separate AUROC analysis.
The equally weighted summary has six endpoint units.  A descriptive hierarchical
bootstrap resamples those tasks and then paired task--seed effects (10,000
replicates); it measures task/algorithmic-seed stability, not laboratory or
held-out-scaffold sampling uncertainty.  Method error reduction uses
$(e_{\mathrm{ERM}}-e_{\mathrm{method}})/e_{\mathrm{ERM}}$.  Every run stores its
configuration, checkpoint, history and predictions.  All 468 primary runs
completed, producing 5,616 metric records and 120 split manifests, and all 46
graph-network control runs completed as well.  All preprocessing, novelty coordinates, thresholds and early
stopping use no test labels.  Unit tests assert split non-overlap and objective
differentiability.
For the secondary worst-view-bin diagnostic, classification error is Brier loss
and regression error is absolute error divided by the training-target standard
deviation; bins with fewer than ten test molecules are omitted before taking the
maximum.  This diagnostic is distinct from Proposition~\ref{prop:control}.

\section*{Acknowledgements}
The authors acknowledge the Comprehensive Marine Natural Products Database (CMNPD) for providing access to the marine natural product data used in this study. This research received no external funding.

\section*{Author contributions}
J.Z. conceived and designed the study, performed the computational experiments, analyzed and interpreted the results, and wrote the original manuscript. C.G. contributed to the methodology, interpretation of the results, and review and editing of the manuscript. Z.W., H.C. contributed to data analysis, validation, and manuscript revision. X.L. contributed to visualization. J.Z., as the corresponding author, supervised and coordinated the project, provided scientific guidance, and critically reviewed and revised the manuscript. All authors read and approved the final manuscript.

\section*{Conflict of Interest}
The authors declare no conflict of interest.

\section*{Data Availability Statement}
The public endpoint tables are Therapeutics Data Commons assets obtained from
Harvard Dataverse under its dataset-level CC0~1.0 licence (DOI
10.7910/DVN/21LKWG, version 105.0).\cite{tdc2020data}  \cmnpd\ is available under
CC~BY-NC-SA~4.0 for scholarly non-commercial use.\cite{cmnpdTerms}  All code,
split manifests, per-run predictions, trained checkpoints and figure-generation
scripts that support the findings of this study are provided in the Supporting
Information and will be deposited in a public archive (Zenodo) upon acceptance;
the raw source terms continue to govern redistribution.

\section*{Use of Language Models}
Language-model assistance was used for code debug and prose editing under human-author
responsibility.  No language model is part of the scientific method, predictor,
data generation or statistical analysis; all citations, quantitative claims and
figures were independently verified against primary sources and recorded outputs.

\printbibliography

@article{bemis1996murcko,
  author  = {Bemis, Guy W. and Murcko, Mark A.},
  title   = {The Properties of Known Drugs. 1. Molecular Frameworks},
  journal = {Journal of Medicinal Chemistry},
  year    = {1996},
  volume  = {39},
  number  = {15},
  pages   = {2887--2893},
  doi     = {10.1021/jm9602928}
}

@article{sahigara2012applicability,
  author  = {Sahigara, Faizan and Mansouri, Kamel and Ballabio, Davide and Mauri, Andrea and Consonni, Viviana and Todeschini, Roberto},
  title   = {Comparison of Different Approaches to Define the Applicability Domain of {QSAR} Models},
  journal = {Molecules},
  year    = {2012},
  volume  = {17},
  number  = {5},
  pages   = {4791--4810},
  doi     = {10.3390/molecules17054791}
}

@article{netzeva2005applicability,
  author  = {Netzeva, Tatiana I. and Worth, Andrew P. and Aldenberg, Tom and Benigni, Romualdo and Cronin, Mark T. D. and Gramatica, Paola and Jaworska, Joanna S. and Kahn, Scott and Klopman, Gilles and Marchant, Carol A. and Myatt, Glenn and Nikolova-Jeliazkova, Nina and Patlewicz, Grace Y. and Perkins, Roger and Roberts, David W. and Schultz, Terry W. and Stanton, David T. and van de Sandt, Johannes J. M. and Tong, Weida and Veith, Gilman and Yang, Chihae},
  title   = {Current Status of Methods for Defining the Applicability Domain of (Quantitative) Structure--Activity Relationships: The Report and Recommendations of {ECVAM} Workshop 52},
  journal = {Alternatives to Laboratory Animals},
  year    = {2005},
  volume  = {33},
  number  = {2},
  pages   = {155--173},
  doi     = {10.1177/026119290503300209}
}

@article{sheridan2012domain,
  author  = {Sheridan, Robert P.},
  title   = {Three Useful Dimensions for Domain Applicability in {QSAR} Models Using Random Forest},
  journal = {Journal of Chemical Information and Modeling},
  year    = {2012},
  volume  = {52},
  number  = {3},
  pages   = {814--823},
  doi     = {10.1021/ci300004n}
}

@article{norinder2014conformal,
  author  = {Norinder, Ulf and Carlsson, Lars and Boyer, Scott and Eklund, Martin},
  title   = {Introducing Conformal Prediction in Predictive Modeling: A Transparent and Flexible Alternative to Applicability Domain Determination},
  journal = {Journal of Chemical Information and Modeling},
  year    = {2014},
  volume  = {54},
  number  = {6},
  pages   = {1596--1603},
  doi     = {10.1021/ci5001168}
}

@article{hirschfeld2020uq,
  author  = {Hirschfeld, Lior and Swanson, Kyle and Yang, Kevin and Barzilay, Regina and Coley, Connor W.},
  title   = {Uncertainty Quantification Using Neural Networks for Molecular Property Prediction},
  journal = {Journal of Chemical Information and Modeling},
  year    = {2020},
  volume  = {60},
  number  = {8},
  pages   = {3770--3780},
  doi     = {10.1021/acs.jcim.0c00502}
}

@article{angelopoulos2023conformal,
  author  = {Angelopoulos, Anastasios N. and Bates, Stephen},
  title   = {Conformal Prediction: A Gentle Introduction},
  journal = {Foundations and Trends in Machine Learning},
  year    = {2023},
  volume  = {16},
  number  = {4},
  pages   = {494--591},
  doi     = {10.1561/2200000101}
}

@article{vanTilborg2023systematic,
  author  = {Deng, Jianyuan and Yang, Zhibo and Wang, Hehe and Ojima, Iwao and Samaras, Dimitris and Wang, Fusheng},
  title   = {A systematic study of key elements underlying molecular property prediction},
  journal = {Nature Communications},
  year    = {2023},
  volume  = {14},
  number  = {1},
  pages   = {6395},
  doi     = {10.1038/s41467-023-41948-6}
}

@article{sheridan2023simpd,
  author  = {Landrum, Gregory A. and Beckers, Maximilian and Lanini, Jessica and Schneider, Nadine and Stiefl, Nikolaus and Riniker, Sereina},
  title   = {{SIMPD}: an algorithm for generating simulated time splits for validating machine learning approaches},
  journal = {Journal of Cheminformatics},
  volume  = {15},
  number  = {1},
  pages   = {119},
  year    = {2023},
  doi     = {10.1186/s13321-023-00787-9}
}

@inproceedings{stanley2023lohi,
  author    = {Steshin, Simon},
  title     = {{Lo-Hi}: Practical {ML} Drug Discovery Benchmark},
  booktitle = {Advances in Neural Information Processing Systems},
  volume    = {36},
  pages     = {64526--64554},
  year      = {2023},
  url       = {https://proceedings.neurips.cc/paper_files/paper/2023/file/cb82f1f97ad0ca1d92df852a44a3bd73-Paper-Datasets_and_Benchmarks.pdf}
}

@article{realworldmood2024,
  author  = {Tossou, Prudencio and Wognum, Cas and Craig, Michael and Mary, Hadrien and Noutahi, Emmanuel},
  title   = {Real-World Molecular Out-Of-Distribution: Specification and Investigation},
  journal = {Journal of Chemical Information and Modeling},
  volume  = {64},
  number  = {3},
  pages   = {697--711},
  year    = {2024},
  doi     = {10.1021/acs.jcim.3c01774}
}

@article{datasail2025,
  author  = {Joeres, Roman and Blumenthal, David B. and Kalinina, Olga V.},
  title   = {Data splitting to avoid information leakage with {DataSAIL}},
  journal = {Nature Communications},
  volume  = {16},
  number  = {1},
  pages   = {3337},
  year    = {2025},
  doi     = {10.1038/s41467-025-58606-8}
}

@article{fooladi2025molecularood,
  author  = {Fooladi, Hosein and Vu, Thi Ngoc Lan and Mathea, Miriam and Kirchmair, Johannes},
  title   = {Evaluating Machine Learning Models for Molecular Property Prediction: Performance and Robustness on Out-of-Distribution Data},
  journal = {Journal of Chemical Information and Modeling},
  volume  = {65},
  number  = {19},
  pages   = {9871--9891},
  year    = {2025},
  doi     = {10.1021/acs.jcim.5c00475}
}

@inproceedings{antoniuk2025boom,
  author    = {Antoniuk, Evan and Zaman, Shehtab and Ben-Nun, Tal and Li, Peggy and Diffenderfer, James and Sahin, Busra and Smolenski, Obadiah and Grethel, Everett and Hsu, Tim and Hiszpanski, Anna and Chiu, Kenneth and Kailkhura, Bhavya and Van Essen, Brian},
  title     = {{BOOM}: Benchmarking Out-Of-distribution Molecular Property Predictions of Machine Learning Models},
  booktitle = {Advances in Neural Information Processing Systems},
  volume    = {38},
  year      = {2025},
  url       = {https://proceedings.neurips.cc/paper_files/paper/2025/file/b94263f7f98c9766ea9a09761ddd88ee-Paper-Datasets_and_Benchmarks_Track.pdf}
}

@article{vantilborg2026edge,
  author  = {van Tilborg, Derek and Rossen, Luke and Grisoni, Francesca},
  title   = {Molecular deep learning at the edge of chemical space},
  journal = {Nature Machine Intelligence},
  volume  = {8},
  number  = {4},
  pages   = {575--587},
  year    = {2026},
  doi     = {10.1038/s42256-026-01216-w}
}

@article{boldini2024fingerprints,
  author  = {Boldini, Davide and Ballabio, Davide and Consonni, Viviana and Todeschini, Roberto and Grisoni, Francesca and Sieber, Stephan A.},
  title   = {Effectiveness of molecular fingerprints for exploring the chemical space of natural products},
  journal = {Journal of Cheminformatics},
  volume  = {16},
  number  = {1},
  pages   = {35},
  year    = {2024},
  doi     = {10.1186/s13321-024-00830-3}
}

@article{lyu2021cmnpd,
  author  = {Lyu, Chuanyu and Chen, Tong and Qiang, Bo and Liu, Ningfeng and Wang, Heyu and Zhang, Liangren and Liu, Zhenming},
  title   = {{CMNPD}: a comprehensive marine natural products database towards facilitating drug discovery from the ocean},
  journal = {Nucleic Acids Research},
  year    = {2021},
  volume  = {49},
  number  = {D1},
  pages   = {D509--D515},
  doi     = {10.1093/nar/gkaa763}
}

@misc{cmnpdParameters,
  author       = {{CMNPD}},
  title        = {Parameter Questions},
  year         = {n.d.},
  howpublished = {CMNPD Interface Documentation},
  url          = {https://docs.cmnpd.org/frequently-asked-questions/parameter-questions},
  note         = {Accessed 2026-07-12}
}

@misc{cmnpdDownloads,
  author       = {{CMNPD}},
  title        = {Downloads},
  year         = {n.d.},
  howpublished = {CMNPD Interface Documentation},
  url          = {https://docs.cmnpd.org/downloads},
  note         = {Accessed 2026-07-12}
}

@misc{cmnpdTerms,
  author       = {{CMNPD}},
  title        = {Terms and Conditions},
  year         = {n.d.},
  howpublished = {CMNPD Interface Documentation},
  url          = {https://docs.cmnpd.org/terms-and-conditions},
  note         = {Accessed 2026-07-12}
}

@inproceedings{huang2021tdc,
  author    = {Huang, Kexin and Fu, Tianfan and Gao, Wenhao and Zhao, Yue and Roohani, Yusuf and Leskovec, Jure and Coley, Connor W. and Xiao, Cao and Sun, Jimeng and Zitnik, Marinka},
  title     = {Therapeutics Data Commons: Machine Learning Datasets and Tasks for Drug Discovery and Development},
  booktitle = {Proceedings of the Neural Information Processing Systems Track on Datasets and Benchmarks},
  editor    = {Vanschoren, Joaquin and Yeung, Sai-Kit},
  volume    = {1},
  year      = {2021},
  eprint    = {2102.09548},
  archivePrefix = {arXiv},
  primaryClass  = {cs.LG},
  url       = {https://datasets-benchmarks-proceedings.neurips.cc/paper_files/paper/2021/hash/4c56ff4ce4aaf9573aa5dff913df997a-Abstract-round1.html}
}

@misc{tdc2020data,
  author    = {Huang, Kexin and Fu, Tianfan and Gao, Wenhao and Zhao, Yue and Roohani, Yusuf and Leskovec, Jure and Coley, Connor W. and Xiao, Cao and Sun, Jimeng and Zitnik, Marinka},
  title     = {Therapeutics Data Commons},
  year      = {2020},
  publisher = {Harvard Dataverse},
  version   = {105.0},
  doi       = {10.7910/DVN/21LKWG},
  url       = {https://doi.org/10.7910/DVN/21LKWG},
  note      = {Retrieved 2026-07-11; dataset-level CC0 1.0 waiver}
}

@misc{arjovsky2019irm,
  title         = {Invariant Risk Minimization},
  author        = {Arjovsky, Martin and Bottou, L{\'e}on and Gulrajani, Ishaan and Lopez-Paz, David},
  year          = {2019},
  eprint        = {1907.02893},
  archivePrefix = {arXiv},
  primaryClass  = {stat.ML},
  doi           = {10.48550/arXiv.1907.02893}
}

@inproceedings{sagawa2020groupdro,
  title     = {Distributionally Robust Neural Networks for Group Shifts: On the Importance of Regularization for Worst-Case Generalization},
  author    = {Sagawa, Shiori and Koh, Pang Wei and Hashimoto, Tatsunori B. and Liang, Percy},
  booktitle = {International Conference on Learning Representations},
  year      = {2020},
  url       = {https://openreview.net/forum?id=ryxGuJrFvS}
}

@inproceedings{krueger2021rex,
  title     = {Out-of-Distribution Generalization via Risk Extrapolation ({REx})},
  author    = {Krueger, David and Caballero, Ethan and Jacobsen, Joern-Henrik and Zhang, Amy and Binas, Jonathan and Zhang, Dinghuai and Le Priol, Remi and Courville, Aaron},
  booktitle = {Proceedings of the 38th International Conference on Machine Learning},
  pages     = {5815--5826},
  year      = {2021},
  volume    = {139},
  series    = {Proceedings of Machine Learning Research},
  publisher = {PMLR},
  url       = {https://proceedings.mlr.press/v139/krueger21a.html}
}

@article{rockafellar2000cvar,
  title   = {Optimization of Conditional Value-at-Risk},
  author  = {Rockafellar, R. Tyrrell and Uryasev, Stanislav},
  journal = {The Journal of Risk},
  volume  = {2},
  number  = {3},
  pages   = {21--41},
  year    = {2000},
  doi     = {10.21314/JOR.2000.038}
}

@inproceedings{li2023imold,
  author    = {Zhuang, Xiang and Zhang, Qiang and Ding, Keyan and Bian, Yatao and Wang, Xiao and Lv, Jingsong and Chen, Hongyang and Chen, Huajun},
  title     = {Learning Invariant Molecular Representation in Latent Discrete Space},
  booktitle = {Advances in Neural Information Processing Systems},
  volume    = {36},
  pages     = {78435--78452},
  year      = {2023},
  url       = {https://proceedings.neurips.cc/paper_files/paper/2023/file/f780a86b7145988ac219d49d8e37a58f-Paper-Conference.pdf}
}

@inproceedings{yu2024minimax,
  title     = {Efficient Algorithms for Empirical Group Distributionally Robust Optimization and Beyond},
  author    = {Yu, Dingzhi and Cai, Yunuo and Jiang, Wei and Zhang, Lijun},
  booktitle = {Proceedings of the 41st International Conference on Machine Learning},
  pages     = {57384--57414},
  year      = {2024},
  volume    = {235},
  series    = {Proceedings of Machine Learning Research},
  publisher = {PMLR},
  url       = {https://proceedings.mlr.press/v235/yu24a.html}
}

@inproceedings{partial2024robustness,
  title     = {Achievable Distributional Robustness When the Robust Risk Is Only Partially Identified},
  author    = {Kostin, Julia and Gnecco, Nicola and Yang, Fanny},
  booktitle = {Advances in Neural Information Processing Systems},
  volume    = {37},
  pages     = {83915--83950},
  year      = {2024},
  doi       = {10.52202/079017-2667}
}

@inproceedings{gui2024invariance,
  title     = {Dissecting the Failure of Invariant Learning on Graphs},
  author    = {Wang, Qixun and Wang, Yifei and Wang, Yisen and Ying, Xianghua},
  booktitle = {Advances in Neural Information Processing Systems},
  volume    = {37},
  pages     = {80383--80438},
  year      = {2024},
  doi       = {10.52202/079017-2556}
}

@article{rogers2010ecfp,
  author  = {Rogers, David and Hahn, Mathew},
  title   = {Extended-Connectivity Fingerprints},
  journal = {Journal of Chemical Information and Modeling},
  year    = {2010},
  volume  = {50},
  number  = {5},
  pages   = {742--754},
  doi     = {10.1021/ci100050t}
}

@article{swanson2024admetai,
  author  = {Swanson, Kyle and Walther, Parker and Leitz, Jeremy and Mukherjee, Souhrid and Wu, Joseph C. and Shivnaraine, Rabindra V. and Zou, James},
  title   = {{ADMET-AI}: a machine learning {ADMET} platform for evaluation of large-scale chemical libraries},
  journal = {Bioinformatics},
  year    = {2024},
  volume  = {40},
  number  = {7},
  pages   = {btae416},
  doi     = {10.1093/bioinformatics/btae416}
}

@article{fu2024admetlab,
  author  = {Fu, Li and Shi, Shaohua and Yi, Jiacai and Wang, Ningning and He, Yuanhang and Wu, Zhenxing and Peng, Jinfu and Deng, Youchao and Wang, Wenxuan and Wu, Chengkun and Lyu, Aiping and Zeng, Xiangxiang and Zhao, Wentao and Hou, Tingjun and Cao, Dongsheng},
  title   = {{ADMETlab 3.0}: an updated comprehensive online {ADMET} prediction platform enhanced with broader coverage, improved performance, {API} functionality and decision support},
  journal = {Nucleic Acids Research},
  year    = {2024},
  volume  = {52},
  number  = {W1},
  pages   = {W422--W431},
  doi     = {10.1093/nar/gkae236}
}

@article{martins2012bbb,
  author  = {Martins, Ines Filipa and Teixeira, Ana L. and Pinheiro, Luis and Falcao, Andre O.},
  title   = {A Bayesian Approach to in Silico Blood-Brain Barrier Penetration Modeling},
  journal = {Journal of Chemical Information and Modeling},
  year    = {2012},
  volume  = {52},
  number  = {6},
  pages   = {1686--1697},
  doi     = {10.1021/ci300124c}
}

@article{hou2007hia,
  author  = {Hou, Tingjun and Wang, Junmei and Li, Youyong},
  title   = {{ADME} Evaluation in Drug Discovery. 8. The Prediction of Human Intestinal Absorption by a Support Vector Machine},
  journal = {Journal of Chemical Information and Modeling},
  year    = {2007},
  volume  = {47},
  number  = {6},
  pages   = {2408--2415},
  doi     = {10.1021/ci7002076}
}

@article{veith2009cyp,
  author  = {Veith, Henrike and Southall, Noel and Huang, Ruili and James, Tim and Fayne, Darren and Artemenko, Natalia and Shen, Min and Inglese, James and Austin, Christopher P. and Lloyd, David G. and Auld, Douglas S.},
  title   = {Comprehensive characterization of cytochrome {P450} isozyme selectivity across chemical libraries},
  journal = {Nature Biotechnology},
  year    = {2009},
  volume  = {27},
  number  = {11},
  pages   = {1050--1055},
  doi     = {10.1038/nbt.1581}
}

@misc{wenlock2015ppbr,
  author    = {Wenlock, Mark and Tomkinson, Nicholas},
  title     = {Experimental in vitro {DMPK} and physicochemical data on a set of publicly disclosed compounds},
  year      = {2015},
  publisher = {EMBL-EBI, ChEMBL},
  doi       = {10.6019/CHEMBL3301361},
  note      = {ChEMBL deposited dataset CHEMBL3301361}
}

@article{sorkun2019aqsoldb,
  author  = {Sorkun, Murat Cihan and Khetan, Abhishek and Er, S{\"u}leyman},
  title   = {{AqSolDB}, a curated reference set of aqueous solubility and 2D descriptors for a diverse set of compounds},
  journal = {Scientific Data},
  year    = {2019},
  volume  = {6},
  number  = {1},
  pages   = {143},
  doi     = {10.1038/s41597-019-0151-1}
}

@article{xu2015dili,
  author  = {Xu, Youjun and Dai, Ziwei and Chen, Fangjin and Gao, Shuaishi and Pei, Jianfeng and Lai, Luhua},
  title   = {Deep Learning for Drug-Induced Liver Injury},
  journal = {Journal of Chemical Information and Modeling},
  year    = {2015},
  volume  = {55},
  number  = {10},
  pages   = {2085--2093},
  doi     = {10.1021/acs.jcim.5b00238}
}

\end{document}


\maketitle

\appendix

\renewcommand{\thesection}{S\arabic{section}}
\renewcommand{\thesubsection}{S\arabic{section}.\arabic{subsection}}
\renewcommand{\thesubsubsection}{S\arabic{section}.\arabic{subsection}.\arabic{subsubsection}}
\renewcommand{\thetable}{S\arabic{table}}
\renewcommand{\thefigure}{S\arabic{figure}}

\setcounter{section}{0}
\setcounter{table}{0}
\setcounter{figure}{0}
\section*{Supplementary Materials}
\noindent This Supporting Information provides data-provenance and integrity
diagnostics, the proof of the coverage proposition stated in the Computational
Methods, leave-one-endpoint-out and label-mixture sensitivity analyses, the full
\cmnpd\ teacher-agreement audit, and reproducibility details.  Values are
computed from recorded experimental outputs, and captions mark the displays that
are illustrative.

\section{Data diagnosis and provenance}
\label{app:data}

Table~\ref{tab:datasets} summarizes the harmonized public endpoint tasks and
their primary metrics.

\begin{table}[ht]
  \caption{\textbf{Harmonized public endpoint tasks.}  Primary metrics follow the TDC
  task convention.}
  \label{tab:datasets}
  \centering
  \small
  \resizebox{\linewidth}{!}{\begin{tabular}{lrlll}
\toprule
Endpoint & $n$ & Task & Target / unit & Primary metric \\
\midrule
BBB permeability & 1,941 & Classification & Permeable class & AUROC $\uparrow$ \\
Intestinal absorption & 577 & Classification & Absorbed class & AUROC $\uparrow$ \\
CYP2D6 inhibition & 12,888 & Classification & Inhibitor class & AUPRC $\uparrow$ \\
Plasma protein binding & 1,578 & Regression & Human binding (percentage points) & MAE $\downarrow$ \\
Aqueous solubility & 9,690 & Regression & Aqueous logS ($\log_{10}$ mol l$^{-1}$) & MAE $\downarrow$ \\
Drug-induced liver injury & 475 & Classification & Injury-positive class & AUROC $\uparrow$ \\
\bottomrule
\end{tabular}
}
\end{table}

\cmnpd\ contains 31,561 rows and 23 columns \cite{cmnpdDownloads}, 9,191 non-empty Murcko scaffolds,
5,065 singleton scaffolds, 1,977 acyclic molecules, 1,204 multi-fragment
structures, 1,081 metal-containing structures and 664 structures with non-zero
formal charge.  Raw SMILES are unique, but connectivity canonicalization yields
27,317 structures.  Thirty-five duplicated Standard InChIKey groups disagree on
at least one teacher output.  BBB continuous scores are absent for 14,995 records;
the corresponding level 4 denotes undefined, not a fifth biological class.
Solubility teacher outputs contain 477 values with absolute magnitude above 10,
including 28 above 100.  These observations motivate the harmonization and
teacher-agreement-only policies used in the main text.

The supplied \cmnpd\ file has SHA-256
\texttt{cdccc990\allowbreak60070c9f\allowbreak f8f56b40\allowbreak
a206e79c\allowbreak05ef3c5d\allowbreak6c99fe3e\allowbreak
b759422f\allowbreak c8a75c23}.
Hashes and source identifiers for all six TDC tables are distributed with the
reproduction artifacts.

\section{Proof of the coverage proposition}
\label{app:proof}

\paragraph{Observed-bin coverage.}
Condition on the positive group counts.  For a predictor, view map and binning
rule fixed independently of the evaluation sample, Hoeffding's inequality for
$\bar\ell\in[0,1]$ gives
\[
\Pr\!\left(R_{v,b}>\widehat R_{v,b}
+\sqrt{\frac{\log(2G/\delta)}{2n_{v,b}}}\right)
\leq \frac{\delta}{2G}<\frac{\delta}{G}.
\]
A union bound over the $G$ groups therefore gives simultaneous coverage with
probability at least $1-\delta$.  For a declared target mixture satisfying
$q_b\geq0$, $\sum_bq_b=1$ and $q_b\leq1/k$ in one view,
\[
R_{Q_0}=\sum_bq_bR_{v,b}
\leq\sum_bq_b\widehat U_{v,b}
\leq \frac1k\sum_{b\in\TopK_k(\widehat U_v)}\widehat U_{v,b}\leq T.
\]
The second inequality is the optimum of a linear programme: it assigns mass
$1/k$ to each of the $k$ largest upper values.  Maximizing over the unknown
declared view yields $T$.

\paragraph{Declared extension.}
By definition, an extension-family member has a covered $Q_0$ and satisfies
$R_Q-R_{Q_0}\leq\rho(S^E+\eta)$.  Adding this declared condition to
$R_{Q_0}\leq T^E$ gives $R_Q\leq T^E+\rho(S^E+\eta)$.  The second statement covers
targets within that declared risk difference; targets with changed within-bin
conditionals or a larger risk difference lie outside it.  Both
statements assume a predictor, view map and binning rule fixed independently of
the evaluation sample.

\section{Additional results and sensitivity analyses}
\label{app:additional}

The leave-one-endpoint-out summary and the BBB frontier-score-quintile
diagnostics are reported in Tables~\ref{tab:loo} and \ref{tab:bbbq},
respectively.

\begin{table}[H]
  \caption{\textbf{Leave-one-endpoint-out sensitivity of the matched frontier
  gap.}  Only BBB removal materially changes the summary; the mean is skew-driven
  by a single endpoint while the median is stable.  Intervals use a
  20,000-replicate descriptive task/seed bootstrap, so the ``None'' row reproduces
  the 10,000-replicate headline (130.3\%, 52.1--246.0\%) within resampling noise.}
  \label{tab:loo}
  \centering
  \small
  \begin{tabular}{lccc}
\toprule
Omitted endpoint & Mean inflation (\%) & 95\% stability interval (\%) & Median (\%) \\
\midrule
None (all six)   & 130.3 & 52.4--247.6 & 87.0 \\
BBB              & \textbf{75.9} & 44.3--116.2 & 59.7 \\
HIA              & 133.5 & 45.5--272.7 & 59.7 \\
CYP2D6           & 147.6 & 58.0--281.0 & 114.2 \\
DILI             & 131.8 & 44.3--272.3 & 59.7 \\
PPBR             & 144.4 & 52.9--278.9 & 114.2 \\
Solubility       & 148.4 & 59.5--281.3 & 114.2 \\
\bottomrule
\end{tabular}

\end{table}

\begin{table}[H]
  \caption{\textbf{BBB frontier-test diagnostics by label-free frontier-score
  quintile (ERM ensemble).}  Predicted permeability rises with novelty while the
  true positive fraction falls, producing the cross-stratum ranking inversion.}
  \label{tab:bbbq}
  \centering
  \small
  \resizebox{\linewidth}{!}{\begin{tabular}{lccccc}
\toprule
Frontier-score quintile & $n$ & Positive fraction & Mean prediction & Mean frontier score & Within-quintile AUROC \\
\midrule
Q1 (nearest) & 78  & 0.654 & 0.714 & 0.796 & 0.600 \\
Q2           & 103 & 0.825 & 0.702 & 0.848 & 0.601 \\
Q3           & 52  & 0.423 & 0.714 & 0.873 & 0.652 \\
Q4           & 79  & 0.228 & 0.803 & 0.917 & 0.439 \\
Q5 (farthest)& 77  & 0.143 & 0.914 & 0.971 & 0.519 \\
\bottomrule
\end{tabular}
}
\end{table}

The maximum simultaneous Hoeffding correction ranges from 0.035 for large CYP2D6
groups to 0.202 for small DILI groups.  Quartile environments avoid the vacuity of
singleton-scaffold certificates, but source-bin slopes do not identify
unseen-frontier behaviour.  The initial literal-acyclic frontier allocation placed
37\% of AqSolDB in the test set; those frozen results are retained as a sensitivity
artifact, and the corrected main split clusters only the empty-scaffold group
without reading labels.

Table~\ref{tab:shiftdecomp} decomposes each matched frontier gap into a
label-mixture (prevalence or outcome-histogram) component and a residual, using
exact class-prevalence reweighting for classification and ten pooled-quantile
outcome bins for regression.  Two facts follow.  First, the standardized AUROC
equals the raw frontier AUROC to three decimals for every classification task,
because rank AUROC cannot change under class-constant reweighting; the AUROC gaps
are ranking phenomena.  Second, the residual component remains sizeable for BBB
(Brier residual 0.138) and solubility (MAE residual 0.427), while for HIA and PPBR
the label-mixture component carries most of the gap.  This residual still combines
support shift, different fitted models and finite-sample variation, so it is
descriptive rather than causal: it quantifies how much of each degradation is
arithmetically attributable to the label distribution.

\begin{table}[H]
  \caption{\textbf{Label-mixture versus residual decomposition of the matched
  frontier gap.}  Classification uses exact class-prevalence reweighting;
  regression uses ten pooled-quantile outcome bins.  ``Std.\ frontier'' is the
  frontier metric under the matched label mixture.  AUROC rows carry no mixture
  component by construction (rank invariance).  The residual is descriptive, not a
  causal effect.}
  \label{tab:shiftdecomp}
  \centering
  \scriptsize
  \resizebox{\linewidth}{!}{\begin{tabular}{llccccc}
\toprule
Task & Metric & Matched & Frontier & Std.\ frontier & Mixture comp. & Residual comp. \\
\midrule
BBB        & AUROC (rank) & 0.879 & 0.409 & 0.409 & --    & --    \\
           & AUPRC        & 0.953 & 0.429 & 0.717 & 0.288 & 0.236 \\
           & Brier        & 0.107 & 0.413 & 0.245 & 0.168 & 0.138 \\
HIA        & AUROC (rank) & 0.717 & 0.538 & 0.538 & --    & --    \\
           & AUPRC        & 0.901 & 0.580 & 0.855 & 0.276 & 0.045 \\
DILI       & AUROC (rank) & 0.811 & 0.584 & 0.584 & --    & --    \\
           & AUPRC        & 0.873 & 0.600 & 0.751 & 0.151 & 0.122 \\
CYP2D6     & AUROC (rank) & 0.798 & 0.712 & 0.712 & --    & --    \\
           & AUPRC        & 0.569 & 0.383 & 0.485 & 0.103 & 0.084 \\
PPBR       & MAE          & 11.234 & 17.897 & 12.253 & 5.644 & 1.019 \\
Solubility & MAE          & 1.373 & 1.915 & 1.800 & 0.115 & 0.427 \\
\bottomrule
\end{tabular}
}
\end{table}

\begin{figure}[ht]
  \centering
  \includegraphics[width=\linewidth]{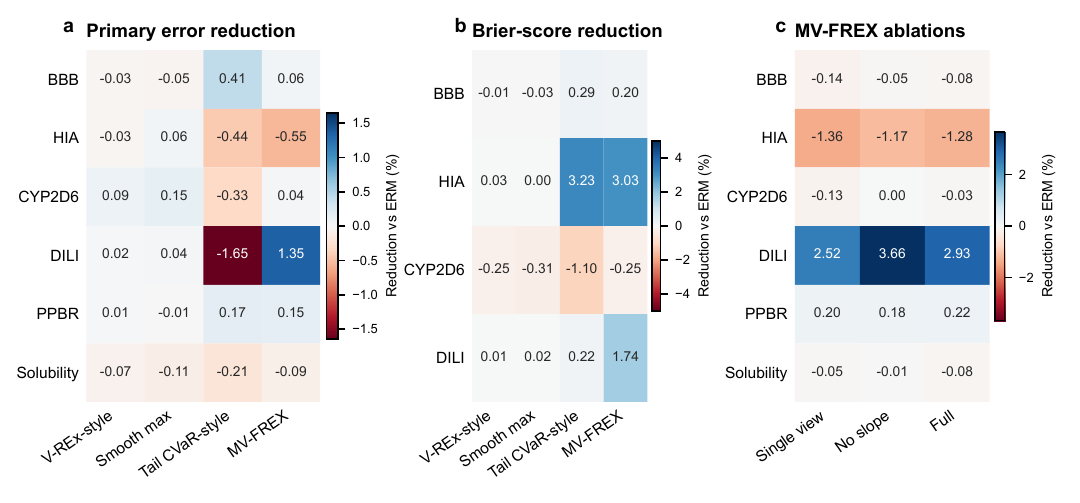}
  \caption{\textbf{Objective effects are small and endpoint-specific.}
  \textbf{a}, Endpoint-wise normalized primary-error reduction relative to ERM.
  \textbf{b}, Brier-score reduction on classification tasks.
  \textbf{c}, Three-seed ablations.  Zero lines mark no change; colour does not
  denote statistical significance.  \emph{Smooth max} and
  \emph{Tail CVaR-style} denote the fixed implementations defined in Experimental
  design, not canonical GroupDRO or a tuned CVaR procedure.}
  \label{fig:ablation}
\end{figure}

Taskwise effects, realized split counts and the three-seed frontier ablation are
reported in Tables~\ref{tab:spliteffects}, \ref{tab:splitcounts} and
\ref{tab:ablations}, respectively; Fig.~\ref{fig:ablation} visualizes the
endpoint-specific objective effects and ablation results.

\begin{table}[H]
  \caption{\textbf{Taskwise matched split effects for ERM.}  Absolute degradation
  is in the primary metric's native scale; chance adjustment is applicable only to
  AUPRC.}
  \label{tab:spliteffects}
  \centering
  \scriptsize
  \resizebox{\linewidth}{!}{\begin{tabular}{llrrr}
\toprule
Task & Metric & Absolute degradation & Error inflation (\%) & Chance-adjusted (\%) \\
\midrule
BBB & AUROC & 0.470 & 402.0 & -- \\
HIA & AUROC & 0.179 & 114.2 & -- \\
CYP2D6 & AUPRC & 0.186 & 43.5 & 31.3 \\
DILI & AUROC & 0.227 & 122.7 & -- \\
PPBR & MAE & 6.663 & 59.7 & -- \\
Solubility & MAE & 0.542 & 39.5 & -- \\
\bottomrule
\end{tabular}
}
\end{table}

\begin{table}[H]
  \caption{\textbf{Realized molecule and group counts.}  The matched scaffold
  control and frontier use the same deterministic acyclic group units and target
  70/10/20 proportions.  Ranges show seed-dependent frontier tie-breaking.}
  \label{tab:splitcounts}
  \centering
  \scriptsize
  \resizebox{\linewidth}{!}{\begin{tabular}{llrrrrrr}
\toprule
& & \multicolumn{2}{c}{Train} & \multicolumn{2}{c}{Validation} & \multicolumn{2}{c}{Test} \\
\cmidrule(lr){3-4}\cmidrule(lr){5-6}\cmidrule(lr){7-8}
Task & Split & Molecules & Groups & Molecules & Groups & Molecules & Groups \\
\midrule
BBB & Matched scaffold & 1359 & 485 & 194 & 194 & 388 & 341 \\
BBB & Structural frontier & 1354--1355 & 686--687 & 197--198 & 127--128 & 389 & 206 \\
HIA & Matched scaffold & 404 & 212 & 58 & 58 & 115 & 115 \\
HIA & Structural frontier & 402 & 253 & 59 & 48 & 116 & 84 \\
CYP2D6 & Matched scaffold & 9021 & 3986 & 1289 & 1289 & 2578 & 2497 \\
CYP2D6 & Structural frontier & 9021 & 5237 & 1289 & 859 & 2578 & 1676 \\
DILI & Matched scaffold & 333 & 169 & 47 & 47 & 95 & 95 \\
DILI & Structural frontier & 316--319 & 206--207 & 48--51 & 37--38 & 108 & 67 \\
PPBR & Matched scaffold & 1104 & 558 & 158 & 158 & 316 & 316 \\
PPBR & Structural frontier & 1103 & 652 & 159 & 130 & 316 & 250 \\
Solubility & Matched scaffold & 6783 & 737 & 969 & 631 & 1938 & 685 \\
Solubility & Structural frontier & 6764--6766 & 1279--1280 & 986--988 & 206--207 & 1938 & 567 \\
\bottomrule
\end{tabular}
}
\end{table}

\begin{table}[ht]
  \caption{\textbf{Three-seed frontier ablation.}  Values are primary metric
  mean $\pm$ standard deviation; metric directions match the primary-results table in the main text.}
  \label{tab:ablations}
  \centering
  \scriptsize
  \resizebox{\linewidth}{!}{\begin{tabular}{lcccc}
\toprule
Task & ERM & Single view & No slope & Full MV-FREX \\
\midrule
BBB & 0.408 $\pm$ 0.006 & 0.408 $\pm$ 0.006 & 0.408 $\pm$ 0.006 & 0.408 $\pm$ 0.005 \\
HIA & 0.533 $\pm$ 0.018 & 0.526 $\pm$ 0.029 & 0.527 $\pm$ 0.029 & 0.527 $\pm$ 0.029 \\
CYP2D6 & 0.383 $\pm$ 0.006 & 0.382 $\pm$ 0.005 & 0.383 $\pm$ 0.006 & 0.383 $\pm$ 0.006 \\
DILI & 0.596 $\pm$ 0.033 & 0.606 $\pm$ 0.034 & 0.610 $\pm$ 0.027 & 0.607 $\pm$ 0.030 \\
PPBR & 17.898 $\pm$ 0.117 & 17.863 $\pm$ 0.113 & 17.867 $\pm$ 0.097 & 17.859 $\pm$ 0.089 \\
Solubility & 1.925 $\pm$ 0.026 & 1.925 $\pm$ 0.012 & 1.925 $\pm$ 0.012 & 1.926 $\pm$ 0.011 \\
\bottomrule
\end{tabular}
}
\end{table}

\clearpage
\section{\cmnpd\ teacher-agreement details}
\label{app:cmnpd}

\begin{figure}[H]
  \centering
  \includegraphics[width=\linewidth]{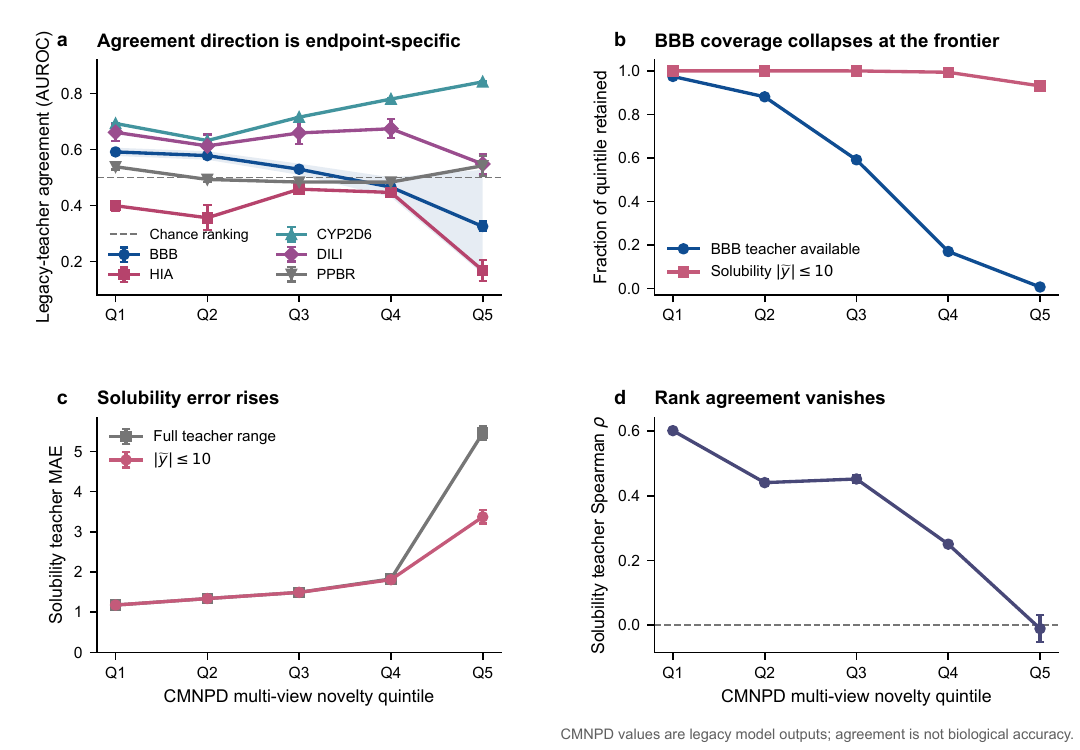}
  \caption{\textbf{Legacy-teacher agreement changes with the multi-view novelty
  frontier.}  \textbf{a}, Classification or threshold AUROC by novelty quintile;
  seed standard deviations are shown, with a compound-bootstrap ribbon for BBB.
  \textbf{b}, Teacher availability and filter retention. \textbf{c}, Full-range
  and $|\widetilde y|\leq10$ solubility teacher MAE. \textbf{d}, Solubility rank
  agreement.  Agreement with a computational teacher is not biological accuracy.}
  \label{fig:teacher}
\end{figure}

The approximate endpoint mappings, overall ensemble agreement and
novelty-stratified agreement are reported in Table~\ref{tab:cmnpdmapping},
Table~\ref{tab:cmnpdagreement} and Fig.~\ref{fig:teacher}, respectively.

\begin{table}[H]
  \caption{\textbf{Approximate mapping from public endpoint tasks to \cmnpd\
  Pipeline Pilot fields.}  Every mapping is used only for teacher agreement.}
  \label{tab:cmnpdmapping}
  \centering
  \scriptsize
  \resizebox{\linewidth}{!}{\begin{tabular}{llll}
\toprule
Task & CMNPD field & Mapping & Scope / handling \\
\midrule
BBB & BBB penetration level & Levels 0--1 vs 2--3 & Level 4 excluded; approximate alignment \\
HIA & HIA level & Level 0 vs levels 1--3 & Stricter teacher threshold; approximate alignment \\
CYP2D6 & CYP2D6 binding model & Binary output & Legacy teacher vs TDC inhibition labels \\
PPBR & Plasma protein binding model & Binary output & TDC regression prediction thresholded at 90\% \\
Solubility & Aqueous solubility & Continuous output & Full and $|\widetilde y|\leq10$ descriptive analyses \\
DILI & Hepatotoxicity model & Binary output & Teacher toxicity vs curated human DILI label \\
\bottomrule
\end{tabular}
}
\end{table}

\begin{table}[H]
  \caption{\textbf{ERM ensemble agreement with \cmnpd\ Pipeline Pilot outputs.}
  These values are not biological accuracy.}
  \label{tab:cmnpdagreement}
  \centering
  \small
  \begin{tabular}{llr}
\toprule
Endpoint mapping & Agreement metric & Value \\
\midrule
BBB & AUROC & 0.565 \\ 
HIA & AUROC & 0.231 \\ 
CYP2D6 & AUROC / AUPRC & 0.733 / 0.134 \\ 
PPBR threshold & AUROC & 0.462 \\ 
Solubility & Spearman / full MAE / filtered MAE & 0.315 / 2.239 / 1.797 \\ 
DILI & AUROC & 0.641 \\ 
\bottomrule
\end{tabular}

\end{table}

\section{Reproducibility, availability and impacts}
\label{app:repro}

\paragraph{Implementation.}
Python 3.13, RDKit 2026.03, scikit-learn 1.7, PyTorch 2.12, NumPy 2.3 and pandas
2.3 were used.  Experiments ran deterministically on a 10-core Apple M4 CPU with
24 GB memory; no CUDA or MPS accelerator was used.  The external hard-split
comparison (main text) used the official Lo-Hi (\texttt{lohi\_splitter})
and DataSAIL (\texttt{datasail-lite}, SCIP backend) packages in an isolated Python
3.11 environment; both splitters were applied to the identical harmonized molecule
sets exported from the main pipeline, and only the resulting index partitions were
returned for training here, so no result depends on the auxiliary environment.  The 468 recorded training runs
required 217.3 aggregate wall-clock seconds after featurization (mean 0.46 s,
maximum 2.06 s per run).  The artifact contains data download, diagnosis,
training, analysis, \cmnpd\ audit, table and Python-only figure scripts, together
with every split manifest and checkpoint.

\printbibliography